\documentclass[letterpaper, 10 pt, conference]{ieeeconf}  

\IEEEoverridecommandlockouts                              

\usepackage{graphicx}                   
\usepackage{mathptmx}                   
\usepackage{times}                      
\usepackage{amsmath}                    
\usepackage{amssymb}                    
\usepackage{algorithm}
\usepackage{algorithmic}
\usepackage{subfig}
\usepackage{balance}
\graphicspath{{figures/}}

\DeclareMathOperator*{\argmax}{argmax}

\newcommand{\comment}[1]{}

\title{\LARGE \bf
A Function Approximation Method for Model-based High-Dimensional Inverse Reinforcement Learning}
\author{Kun Li$^{1}$, Joel W. Burdick$^{1}$
\thanks{*This work was
supported by the National Institutes of Health, NIBIB.} \thanks{$^{1}$Kun Li and Joel W. Burdick are
with Department of Mechanical and Civil Engineering, California Institute of Technology, Pasadena,
CA 91125, USA {\tt\small kunli@caltech.edu}
}%
}
\begin{document}
\maketitle
\thispagestyle{empty}
\pagestyle{empty}
\begin{abstract}
  This works handles the inverse reinforcement learning problem in high-dimensional state spaces,
  which relies on an efficient solution of model-based high-dimensional reinforcement learning
  problems. To solve the computationally expensive reinforcement learning problems, we
  propose a function approximation method to ensure that the Bellman Optimality Equation always
  holds, and then estimate a function based on the observed human actions for inverse reinforcement
  learning problems. The time complexity of the proposed method is linearly proportional to the
  cardinality of the action set, thus it can handle high-dimensional even continuous state spaces
  efficiently. We test the proposed method in a simulated environment to show its accuracy,
  and  three clinical tasks to show how it can be used to evaluate a doctor's proficiency.
\end{abstract}
\section{Introduction}
\label{irl::intro}
Recently, surgical robots, like Da Vinci Surgical System, have been applied to many tasks, due to
its reliability and accuracy. In these systems, a doctor operates the robot manipulator remotely,
and gets the visual feedback during a surgery. With a sophisticated control system and
high-resolution images, the surgery can be done with higher precision and less accidents.
However, this requires the doctor to concentrate on robot operations and visual feedbacks during the
whole surgery, which may lead to fatigue and errors. 

\begin{figure}
  \centering
  \includegraphics[width=0.4\textwidth]{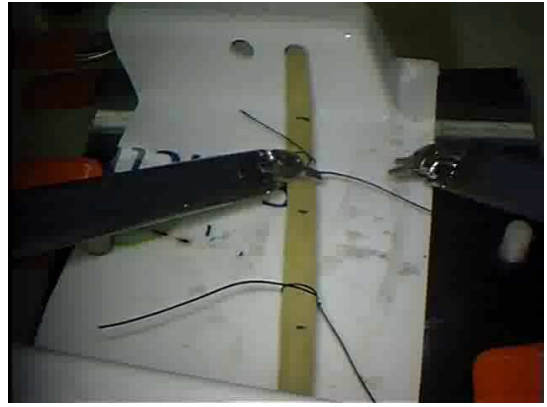}
  \caption{Knot tying with Da Vinci robot: the photo is grabbed from JIGSAW dataset
  \cite{irl::jigsaw}.}
  \label{fig:surgicalrobot}
\end{figure}

To solve the problem, some level of automation can be introduced, considering that many
surgeries contain repeating atomic operations. For example, knot tying is a typical procedure after
many surgeries, as shown in Figure \ref{fig:surgicalrobot}, and it can be decomposed into a sequence
of pre-trained standard operations for the robot. The automation can also be used to avoid possible
mistakes committed by an inexperienced doctor during a surgery, where alarm signal can be triggered
when an unusual  action is taken by the doctor, and the amount of alarm signals can be used to
evaluate the doctor as well.

The core of the automation system is a control policy, predicting which action to take under each
state for typical surgical robots. The control policy can be defined manually, but it is difficult
due to the possible number of states occurring during a surgery. Another solution is estimating the
policy by solving a Markov decision process, but it needs an accurate reward function, depending on
too many factors to be defined manually.

An alternative solution is learning the control policy from experts' demonstrations through
imitation learning. Many algorithms try to learn the policy from the state-action pair directly
in a supervised way, but the learned policy usually does not indicate how good a state-action pair
is, which is useful for online doctor action evaluation. This problem can be solved by inverse
reinforcement learning algorithms, which learns a reward function from the observed demonstrations,
and the optimality of a control policy can be estimated based on the reward function.

Existing solutions of the inverse reinforcement learning problem mainly work on small-scale
problems, by collecting a set of observations for reward estimation and using the estimated reward
afterwards.  For example, the methods in \cite{irl::irl1,irl::irl2, irl::subgradient} estimate the
agent's policy from a set of observations, and estimate a reward function that leads to the policy.
The method in \cite{irl::maxentropy} collects a set of trajectories of the agent, and estimates a
reward function that maximizes the likelihood of the trajectories. This strategy works for
applications in small state spaces. However, the state space of sensory feedback is huge for
surgical evaluation, and these method cannot handle it well due to the reinforcement learning
problem in each iteration of reward estimation.

Some existing methods can be scaled to high-dimensional state spaces and solve the problem without
learning the transition model. While they improve the learning efficiency, they cannot utilize
unsupervised data, or data from the demonstrations of non-experts. These data cannot be used to
learn the reward function, but they provide information about the environment dynamics.

In this work, we find that inverse reinforcement learning in high-dimensional space can be
simplified under the condition that the transition model and the set of action remain unchanged for
the subject, where each reward function leads to a unique optimal value function. Based on this
assumption, we propose a function approximation method that learns the reward function and the
optimal value function, but without the computationally expensive reinforcement learning steps, thus
it can be scaled to high dimensional state spaces. This method can also solve  model-based
high-dimensional reinforcement learning problems, although it is not our main focus. 

The paper is organized as follows. We review existing work on inverse reinforcement learning in
Section \ref{irl::related}, and formulate the function approximation inverse reinforcement learning
method for high-dimensional  problems in \ref{irl::largeirl}. A simulated experiment and a clinical
experiment are shown in Section \ref{irl::experiments}, with conclusions in Section
\ref{irl::conclusions}.

\section{Related Works}
\label{irl::related}
Approximate dynamic programming for reinforcement learning is a well-researched topic in Markov
decision process. A good introduction is given in \cite{irl::approximatedynamic}. Some model-free
methods produce many promising results in recent years, like deep Q network \cite{irl::deepQ},
double Q learning \cite{irl::doubleDeepQ}, advantage learning \cite{irl::advantage}, etc. But in
many robotic applications, reward values are not available for all robot actions, and those data is
wasted in model-free learning. Common model-based approximation methods use a function to
approximate the value function or the Q function, and the performance depends on the selected
features.  

Inverse Reinforcement Learning problem is firstly formulated in \cite{irl::irl1}, where the agent
observes the states resulting from an assumingly optimal policy, and tries to learn a reward
function that makes the policy better than all alternatives. Since the goal can be achieved by
multiple reward functions, this paper tries to find one that maximizes the difference between the
observed policy and the second best policy. This idea is extended by \cite{irl::maxmargin}, in the
name of max-margin learning for inverse optimal control. Another extension is proposed in
\cite{irl::irl2}, where the purpose is not to recover the real reward function, but to find a
reward function that leads to a policy equivalent to the observed one, measured by the amount of
rewards collected by following that policy.

Since a motion policy may be difficult to estimate from observations, a behavior-based method is
proposed in \cite{irl::maxentropy}, which models the distribution of behaviors as a maximum-entropy
model on the amount of reward collected from each behavior. This model has many applications and
extensions. For example, \cite{irl::sequence} considers a sequence of changing reward functions
instead of a single reward function. \cite{irl::gaussianirl} and \cite{irl::guidedirl} consider
complex reward functions, instead of linear one, and use Gaussian process and neural networks,
respectively, to model the reward function. \cite{irl::pomdp} considers complex environments,
instead of a well-observed Markov Decision Process, and combines partially observed Markov Decision
Process with reward learning. \cite{irl::localirl} models the behaviors based on the local
optimality of a behavior, instead of the summation of rewards.  \cite{irl::deepirl} uses a
multi-layer neural network to represent nonlinear reward functions.

Another method is proposed in \cite{irl::bayirl}, which models the probability of a behavior as the
product of each state-action's probability, and learns the reward function via maximum a posteriori
estimation. However, due to the complex relation between the reward function and the behavior
distribution, the author uses computationally expensive Monte-Carlo methods to sample the
distribution. This work is extended by \cite{irl::subgradient}, which uses sub-gradient methods to
simplify the problem.  Another  extensions is shown in \cite{irl::bayioc}, which tries to find a
reward function that matches the observed behavior. For motions involving multiple tasks and varying
reward functions, methods are developed in \cite{irl::multirl1} and \cite{irl::multirl2}, which try
to learn multiple reward functions. 

Most of these methods need to solve a reinforcement learning problem in each step of reward
learning, thus practical large-scale application is computationally infeasible. Several methods are
applicable to large-scale applications. The method in \cite{irl::irl1} uses a linear approximation
of the value function, but it requires a set of manually defined basis functions. The methods in
\cite{irl::guidedirl,irl::relative} update the reward function parameter by minimizing the relative
entropy between the observed trajectories and a set of sampled trajectories based on the reward
function, but they require a set of manually segmented trajectories of human motion, where the
choice of trajectory length will affect the result. The method in \cite{irl::value} only learns an
optimal value function, instead of the reward function.

\section{High-dimensional Inverse Reinforcement Learning}
\label{irl::largeirl}
\subsection{Markov Decision Process}
A Markov Decision Process is described with the following variables:
\begin{itemize}
  \item $S=\{s\}$, a set of states
  \item $A=\{a\}$, a set of actions
  \item $P_{ss'}^a$, a state transition function that defines the probability that state $s$ becomes
    $s'$ after action $a$.
  \item $R=\{r(s)\}$, a reward function that defines the immediate reward of state $s$.
  \item $\gamma$, a discount factor that ensures the convergence of the MDP over an infinite
    horizon.
\end{itemize}

An agent's  motion can be represented as a sequence of state-action pairs:
\[\zeta=\{(s_i,a_i)|i=0,\cdots,N_\zeta\},\]
where $N_\zeta$ denotes the length of the motion, varying in different observations. Given the
observed sequence, inverse reinforcement learning algorithms try to recover a reward function that
explains the motion.

One key problem is how to model the action in each state, or the policy, $\pi(s)\in A$, a mapping
from states to actions. This problem can be handled by reinforcement learning algorithms, by
introducing the value function $V(s)$ and the Q-function $Q(s,a)$, described by the Bellman Equation
\cite{irl::rl}:
\begin{align}
  &V^\pi(s)=\sum_{s'|s,\pi(s)}P_{ss'}^{\pi(s)}[r(s')+\gamma*V^\pi(s')],\\
  &Q^\pi(s,a)=\sum_{s'|s,a}P_{ss'}^a[r(s')+\gamma*V^\pi(s')],
\end{align}
where $V^\pi$ and $Q^\pi$ define the value function and the Q-function under a policy $\pi$.

For an optimal policy $\pi^*$, the value function and the Q-function should be maximized on every
state. This is described by the Bellman Optimality Equation \cite{irl::rl}:
\begin{align}
  &V^*(s)=\max_{a\in A}\sum_{s'|s,a}P_{ss'}^a[r(s')+\gamma*V^*(s')],\\
  &Q^*(s,a)=\sum_{s'|s,a}P_{ss'}^a[r(s')+\gamma*\max_{a'\in A}Q^*(s',a')].
\end{align}

In typical inverse reinforcement learning algorithms, the Bellman Optimality Equation needs to be
solved once for each parameter updating of the reward function, thus it is computationally
infeasible in high-dimensional state spaces. While several existing approaches solve the problem at
the expense of the optimality, we propose an approximation method to avoid the problem.

\subsection{Function Approximation Framework}
Given the set of actions and the transition probability, a reward function leads to a unique optimal
value function. To learn the reward function from the observed motion, instead of directly
learning the reward function, we use a parameterized function, named as \textit{VR function}, to
represent the summation of the reward function and the discounted value function:
\begin{equation}
  f(s,\theta)=r(s)+\gamma*V^*(s).
  \label{equation:approxrewardvalue}
\end{equation}
The function value of a state is named as \textit{VR value}.

Substituting Equation \eqref{equation:approxrewardvalue} into Bellman Optimality Equation, the
optimal Q function is given as:
\begin{equation}
  Q^*(s,a)=\sum_{s'|s,a}P_{ss'}^af(s',\theta),
  \label{equation:approxQ}
\end{equation}
the optimal value function is given as:
\begin{align}
  V^*(s)&=\max_{a\in A}Q^*(s,a)\nonumber\\
        &=\max_{a\in A}\sum_{s'|s,a}P_{ss'}^af(s',\theta),
  \label{equation:approxV}
\end{align}
and the reward function can be computed as:
\begin{align}
  r(s)&=f(s,\theta)-\gamma*V^*(s)\nonumber\\
      &=f(s,\theta)-\gamma*\max_{a\in A}\sum_{s'|s,a}P_{ss'}^af(s',\theta).
  \label{equation:approxR}
\end{align}

Note that this formulation can be generalized to other extensions of Bellman Optimality Equation by
replacing the $max$ operator with other types of Bellman backup operators. For example,
$V^*(s)=\log_{a\in A}\exp Q^*(s,a)$ is used in the maximum-entropy method\cite{irl::maxentropy};
$V^*(s)=\frac{1}{k}\log_{a\in A}\exp k*Q^*(s,a)$ is used in Bellman Gradient Iteration
\cite{irl::BGI}.  

For any \textit{VR function} $f$ and any parameter $\theta$, the optimal Q function $Q^*(s,a)$,
optimal value function $V^*(s)$, and reward function $r(s)$ constructed with Equation
\eqref{equation:approxQ}, \eqref{equation:approxV}, and \eqref{equation:approxR} always meet the
Bellman Optimality Equation. Under this condition, we try to recover a parameterized function
$f(s,\theta)$ that best explains the observed rewards for reinforcement learning problems, and
expert demonstrations $\zeta$ for inverse reinforcement learning problems.

For reinforcement learning problems, the Bellman backup operator should be a differentiable one,
thus the function parameter can be updated based on the observed rewards.

For inverse reinforcement learning problems, combined with different Bellman backup operators, this
formulation can extend many existing methods to high-dimensional space, like the motion model in
\cite{irl::motionvalue}, $p(a|s)=-v^*(s)-\log\sum_k p_{s,k}\exp(-v^*(k))$, the motion model in
\cite{irl::maxentropy}, $p(a|s)=\exp{Q^*(s,a)-V^*(s)}$, and the motion model in \cite{irl::bayirl},
$p(a|s)\propto \exp{Q^*(s,a)}$. The main limitation is the assumption of a known transition model
$P_{ss'}^a$, but it only requires a partial model on the visited states rather than a full
environment model, and it can be learned independently in an unsupervised way.

\subsection{High-dimensional Reinforcement Learning}
Although it is not our main focus, we briefly show how the proposed method solves high-dimensional
reinforcement learning problems. Assuming the approximation function is a neural network, the
parameter $\theta=\{w,b\}$-weights and biases-in Equation \eqref{equation:approxrewardvalue} can be
estimated from the observed sequence of rewards $\hat{R_s}$ via least-square estimation, where the
objective function is:
\[LSE(\theta)=\sum_{s}||\hat{R_s}-r(s)||^2.\]

The reward function $r(s)$ in Equation \eqref{equation:approxR} is non-differentiable with the max
function as the Bellman backup operator. By approximating it with the generalized softmax function
\cite{irl::BGI}, the gradient of the objective function is: \[\nabla_\theta
LSE(\theta)=\sum_{s}2*||\hat{R_s}-r(s)||*(-\nabla_\theta
r(s)),\]
where
\[\nabla_\theta r(s)=\nabla_\theta f(s,\theta)-\gamma*\sum_{a\in
A}\frac{\exp(kQ^*(s,a))}{\sum_{a'\in A}\exp(kQ^*(s,a))}\sum_{s'|s,a}P_{ss'}^a \nabla_\theta
f(s,\theta),\]
 and $k$ is the approximation level.

The parameter $\theta$ can be learned with gradient methods. The algorithm is shown in Algorithm
\ref{alg:nnapproxrl}. With the learned parameter, the optimal value function and a control policy
can be estimated.

\begin{algorithm}[tb]
  \caption{Function Approximation RL with Neural Network}
  \label{alg:nnapproxrl}
\begin{algorithmic}[1]
  \STATE Data: {$R,S,A,P,\gamma,b,\alpha$}
  \STATE Result: {optimal value $V^*[S]$, optimal action value $Q^*[S,A]$}
  \STATE create variable $\theta=\{W,b\}$ for a neural network
  \STATE build $f[S,\theta]$ as the output of the neural network
  \STATE build $Q^*[S,A]$, $V^*[S]$, and $R[S]$ based on Equation \eqref{equation:approxrewardvalue},
  \eqref{equation:approxQ}, \eqref{equation:approxV}, and \eqref{equation:approxR}.
  \STATE build objective function $LSE[\theta]$ based on $R[S]$
  \STATE compute gradient $\nabla_\theta LSE[\theta]$
  \STATE initialize $\theta$
  \WHILE{not converging}
    \STATE $\theta=\theta+\alpha*\nabla_\theta LSE[\theta]$
  \ENDWHILE
  \STATE evaluate {optimal value $V^*[S]$, optimal action value $Q^*[S,A]$}
  \STATE return $Q^*[S,A]$
\end{algorithmic}
\end{algorithm}
\subsection{High-dimensional Inverse Reinforcement Learning}
For IRL problems, this work chooses $max$ as the Bellman backup operator
and a motion model $p(a|s)$ based on the optimal Q function $Q^*(s,a)$ \cite{irl::bayirl}:
\begin{equation}
  P(a|s)=\frac{\exp{b*Q^*(s,a)}}{\sum_{\tilde{a}\in
  A}\exp{b*Q^*(s,\tilde{a})}},
  \label{equation:motionmodel}
\end{equation} 
where $b$ is a parameter controlling the degree of confidence in the agent's ability to choose
actions based on Q values. In the remaining sections, we use $Q^*(s,a)$ to denote the optimal Q values
for simplified notations.

Assuming the approximation function is a neural network, the parameter $\theta=\{w,b\}$-weights and
biases-in Equation \eqref{equation:approxrewardvalue} can be estimated from the observed sequence
of state-action pairs $\zeta$ via maximum-likelihood estimation:
\begin{equation}
  \label{equation:maxtheta}
  \theta=\argmax_{\theta}\log{P(\zeta|\theta)},
\end{equation}
where the log-likelihood of $P(\zeta|\theta)$ is given by:
\begin{align}
  L(\theta)&=\log{P(\zeta|\theta)}\nonumber\\
           &=\log{\prod_{(s,a)\in \zeta} P(a|\theta;s)}\nonumber\\
           &=\log{\prod_{(s,a)\in \zeta} \frac{\exp{b*Q^*(s,a)}}{\sum_{\hat{a}\in A}\exp{b*Q^*(s,\hat{a})}}
}\nonumber\\
  &=\sum_{(s,a)\in\zeta}(b*Q^*(s,a)-\log{\sum_{\hat{a}\in
  A}\exp{b*Q^*(s,\hat{a}))}},
  \label{equation:loglikelihood}
\end{align}
and the gradient of the log-likelihood is given by:
\begin{align}
  \nabla_\theta L(\theta)&=\sum_{(s,a)\in\zeta}(b*\nabla_\theta Q^*(s,a)\nonumber\\
  &-b*\sum_{\hat{a}\in
  A}P((s,\hat{a})|r(\theta))\nabla_\theta Q^*(s,\hat{a})).
  \label{equation:loglikelihoodgradient}
\end{align}

With a differentiable approximation function, 
\[\nabla_\theta Q^*(s,a)=\sum_{s'|s,a}P_{ss'}^a\nabla_\theta f(s',\theta),\]
and 
\begin{align}
  \nabla_\theta L(\theta)&=\sum_{(s,a)\in\zeta}(b*\sum_{s'|s,a}P_{ss'}^a\nabla_\theta f(s',\theta)\nonumber\\
  &-b*\sum_{\hat{a}\in
  A}P((s,\hat{a})|r(\theta))\sum_{s'|s,a}P_{ss'}^a\nabla_\theta f(s',\theta)),
  \label{equation:loglikelihoodgradient}
\end{align}
where $\nabla_\theta f(s',\theta)$ denotes the gradient of the neural network output with respect to neural
network parameter $\theta=\{w,b\}$.

If the \textit{VR function} $f(s,\theta)$ is linear, the objective function in Equation
\eqref{equation:loglikelihood} is concave, and a global optimum exists. However, a multi-layer
neural network works better to handle the non-linearity in approximation and the high-dimensional
state space data. 

A gradient ascent method is used to learn the parameter $\theta$:
\begin{equation}
  \label{equation:gradientascent}
  \theta=\theta+\alpha*\nabla_\theta L(\theta),
\end{equation}
where $\alpha$ is the learning rate.

When the method converges, we can compute the optimal Q function, the optimal value function, and the
reward function based on Equation \eqref{equation:approxrewardvalue}, \eqref{equation:approxQ},
\eqref{equation:approxV}, and \eqref{equation:approxR}. The algorithm under a neural network-based
approximation function is shown in Algorithm \ref{alg:nnapprox}.

This method does not involve solving the MDP problem for each updated parameter $\theta$, and
large-scale state space can be easily handled by an approximation function based on a
multi-layer neural network. 

Obviously, the approximation function is not unique, but all of them will generate the same optimal
values and rewards for the observed state-action pairs after convergence. By choosing a neural
network with higher capacity, we may overfit the observed state-action distribution, and do not
generalize well. Therefore, the choice of the approximation function depends on how well the observed
motion matches the ground truth one.
\begin{algorithm}[tb]
  \caption{Function Approximation IRL with Neural Network}
  \label{alg:nnapprox}
\begin{algorithmic}[1]
  \STATE Data: {$\zeta,S,A,P,\gamma,b,\alpha$}
  \STATE Result: {optimal value $V^*[S]$, optimal action value $Q^*[S,A]$, reward value $R[S]$}
  \STATE create variable $\theta=\{W,b\}$ for a neural network
  \STATE build $f[S,\theta]$ as the output of the neural network
  \STATE build $Q^*[S,A]$, $V^*[S]$, and $R[S]$ based on Equation \eqref{equation:approxrewardvalue},
  \eqref{equation:approxQ}, \eqref{equation:approxV}, and \eqref{equation:approxR}.
  \STATE build loglikelihood $L[\theta]$ based on $\zeta$ and $Q^*[S,A]$
  \STATE compute gradient $\nabla_\theta L[\theta]$
  \STATE initialize $\theta$
  \WHILE{not converging}
    \STATE $\theta=\theta+\alpha*\nabla_\theta L[\theta]$
  \ENDWHILE
  \STATE evaluate {optimal value $V^*[S]$, optimal action value $Q^*[S,A]$, reward value $R[S]$}
  \STATE return $R[S]$
\end{algorithmic}
\end{algorithm}

\section{Experiments}
\label{irl::experiments}
We first test the proposed method in a simulated environment, to compare its accuracy under
different approximation functions, and then apply the proposed method to surgical data in JIGSAW
dataset \cite{irl::jigsaw}.
\subsection{Simulated Environment}
We create a four-dimensional grids, with 10 grid in each dimension, thus 10000 states are generated.
Several reward-emitting objects are put randomly in the grid, and each of them generates an
exponentially decaying negative or positive reward value to all the grid based on the distances. The
true reward value of each grid is the summation of the generated rewards in the grid. An agent moves
in the grids, and it can choose to move up, down, or stay still in each dimension, described by an
action set of $3^4=81$ actions. The observable feature of a grid is the grid's distances to the
reward-generating objects.

To test the application of the proposed method to reinforcement learning problems, we assume that
the reward value of each state is available for the robot, and it has to learn an optimal value
function from it. We compare the ground truth value function, computed through value iteration,  and
the value function recovered by the robot based on the mean error of the optimal Q values. 

\begin{figure}
  \centering
  \includegraphics[width=0.4\textwidth]{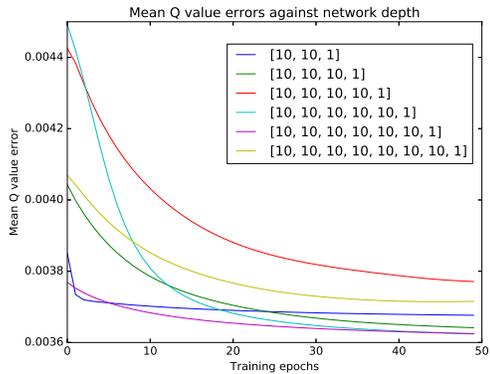}
  \caption{The error of learned Q values versus network depth in RL: the compared networks range from
  three layers to eight layers, including the input layer and the output layers. Under each network
  configuration, a Q function is learned based on the observed reward values, and the ground Q
  function is computed via value iteration. The mean error of the Q values is computed and plotted
  under different number of gradient iterations.}
  \label{fig:qdepth}
\end{figure}

\begin{figure}
  \centering
  \includegraphics[width=0.4\textwidth]{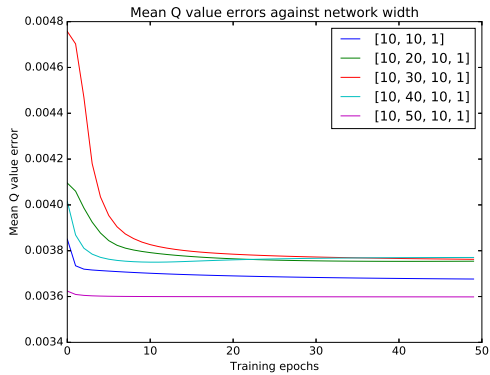}
  \caption{The error of learned Q values versus network width in RL: the hidden nodes of the compared
  networks range from ten to fifty, in addition to an input layer, a common hidden layer,  and a
  output layers. Under each network configuration, a Q function is learned based on the observed
  reward values, and the ground Q function is computed via value iteration. The mean error of the Q
  values is computed and plotted under different number of gradient iterations.}
  \label{fig:qwidth}
\end{figure}

We choose neural networks as the approximation function, and compare the errors under different
neural net configurations. We choose the configuration by firstly fixing the number of nodes in each
hidden layer and increasing the number of layers, and then fixing the number of hidden layers and
increasing the number of nodes in each layer. Stochastic gradient descent is used in optimization,
with batch size 50 and learning rate 0.00001. The result is shown in Figure \ref{fig:qdepth} and
\ref{fig:qwidth}.

\begin{figure}
  \centering
  \includegraphics[width=0.4\textwidth]{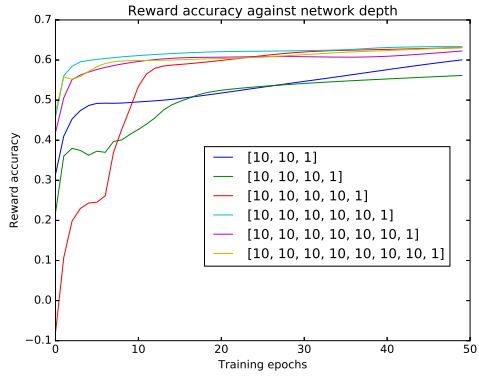}
  \caption{The accuracy of learned reward values versus network depth in IRL: the compared networks range
  from three layers to eight layers, including the input layer and the output layers.  Under each
  network configuration, a reward function is learned based on the observed actions, and the
  accuracy is computed as the correlation coefficient between the learned reward and the ground
  truth reward.}
  \label{fig:rdepth}
\end{figure}

\begin{figure}
  \centering
  \includegraphics[width=0.4\textwidth]{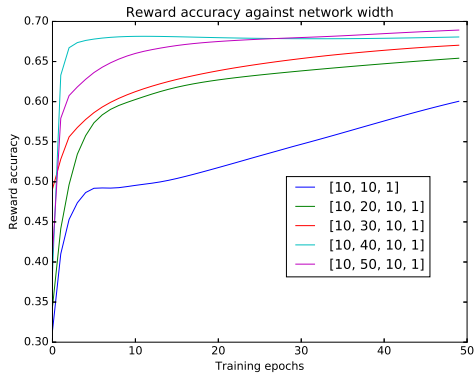}
  \caption{The accuracy of learned reward values  versus network width in IRL: the hidden nodes of the compared
  networks range from ten to fifty, in addition to an input layer, a common hidden layer,  and a
  output layers. Under each network configuration, a reward function is learned based on the observed
  actions, and the accuracy is computed as the correlation coefficient between the learned reward
  and the ground truth reward.}
  \label{fig:rwidth}
\end{figure}

To test the application of the proposed method to inverse reinforcement learning problems, we
generate $200000$ trajectories with random initial position and length $10$ based on the true reward
function, and try to recover a reward function based on the trajectories. We compute the accuracy
based on the correlation coefficient between the ground truth reward function and the recovered
reward function. Similarly, we compare the accuracy under different neural network configurations.
The result is shown in Figure \ref{fig:rdepth} and \ref{fig:rwidth}.

The results show that the accuracies of learned value function and reward function improve as the
capacity of network increases, and increasing network width works better.

\subsection{Surgical Robot Operator}
\begin{figure}
  \centering
  \subfloat[Knot tying]{\includegraphics[width=0.15\textwidth]{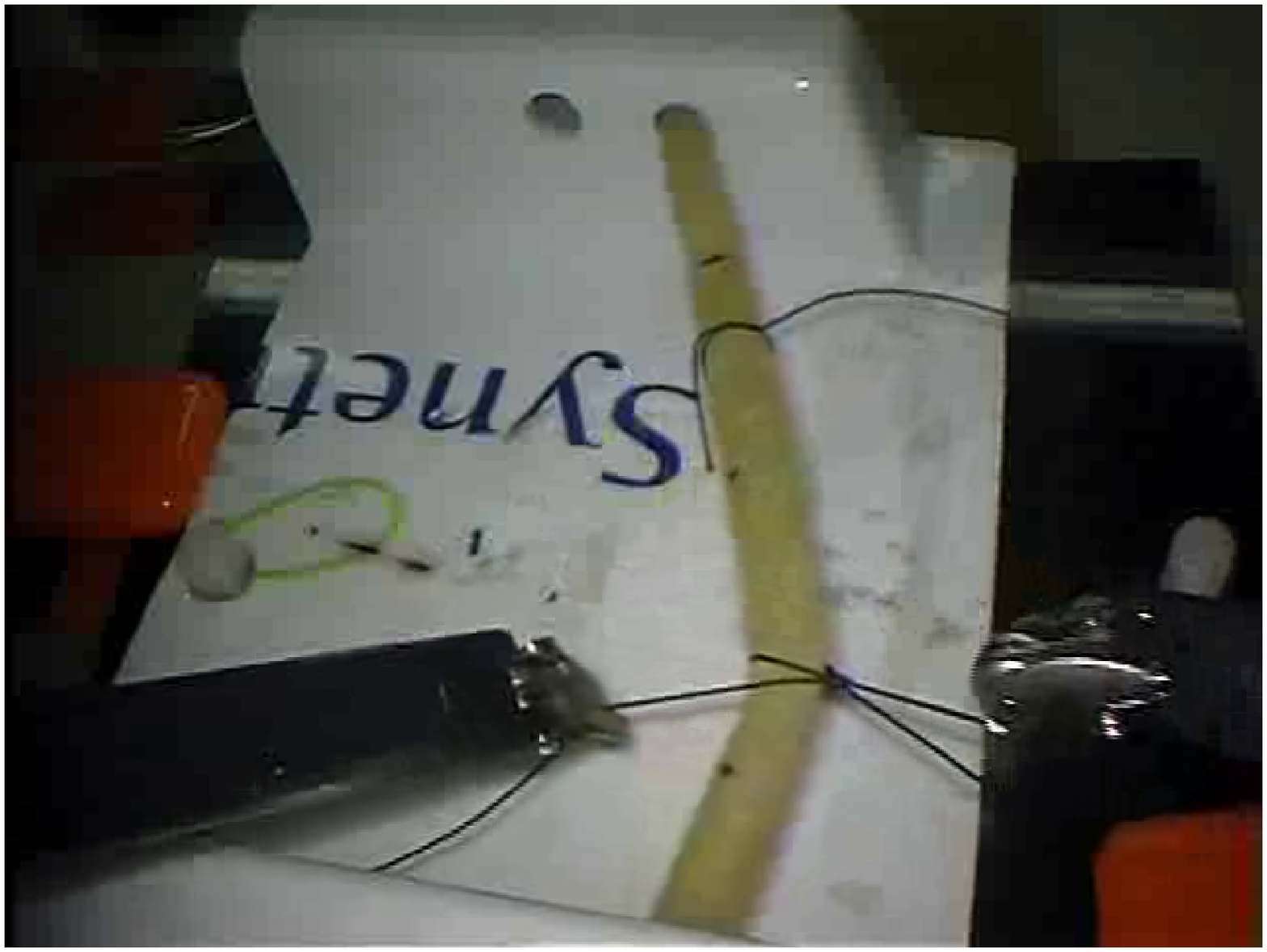}}
  ~ 
  \subfloat[Needle passing]{\includegraphics[width=0.15\textwidth]{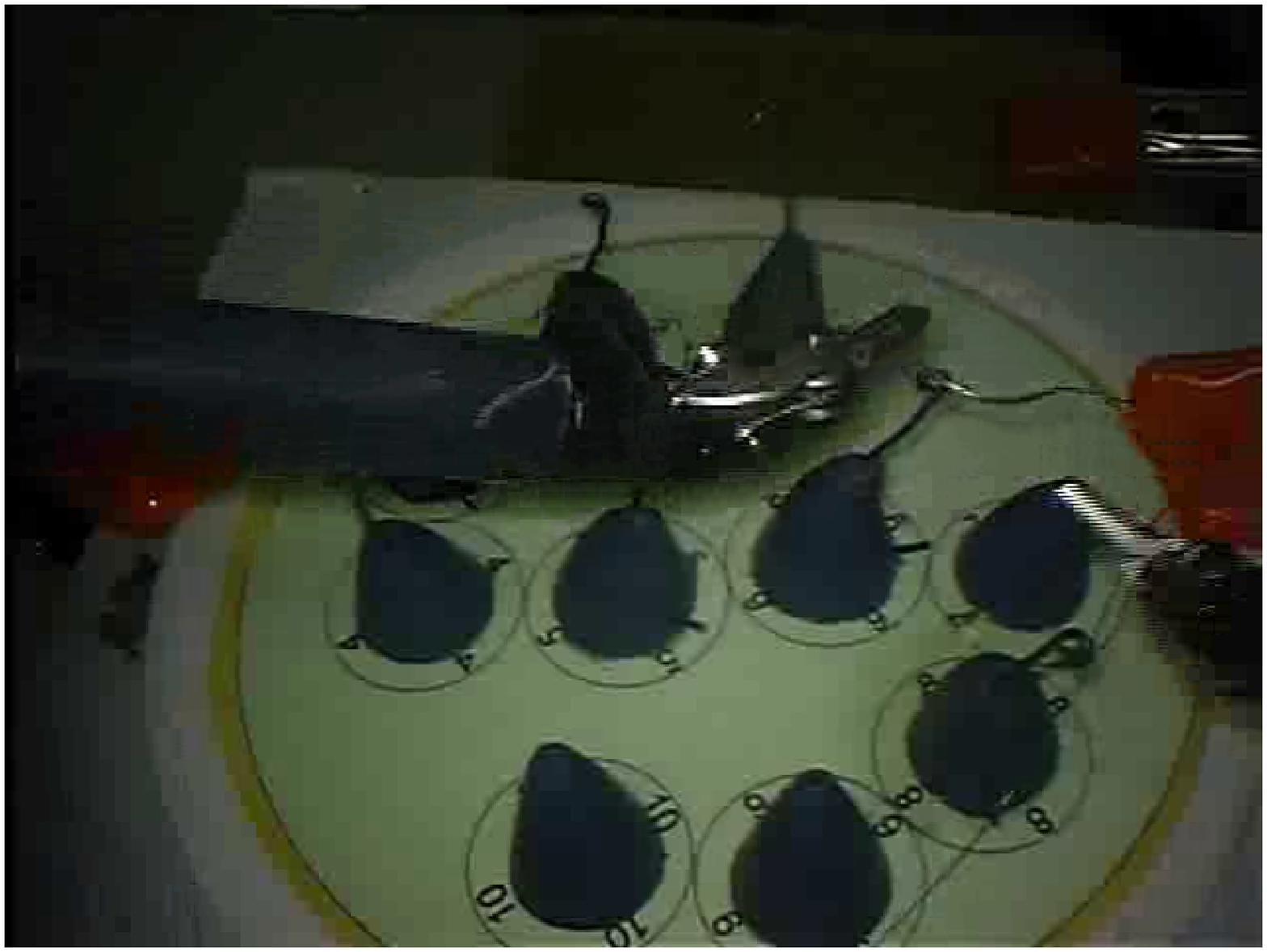}}
  ~
  \subfloat[Suturing]{\includegraphics[width=0.15\textwidth]{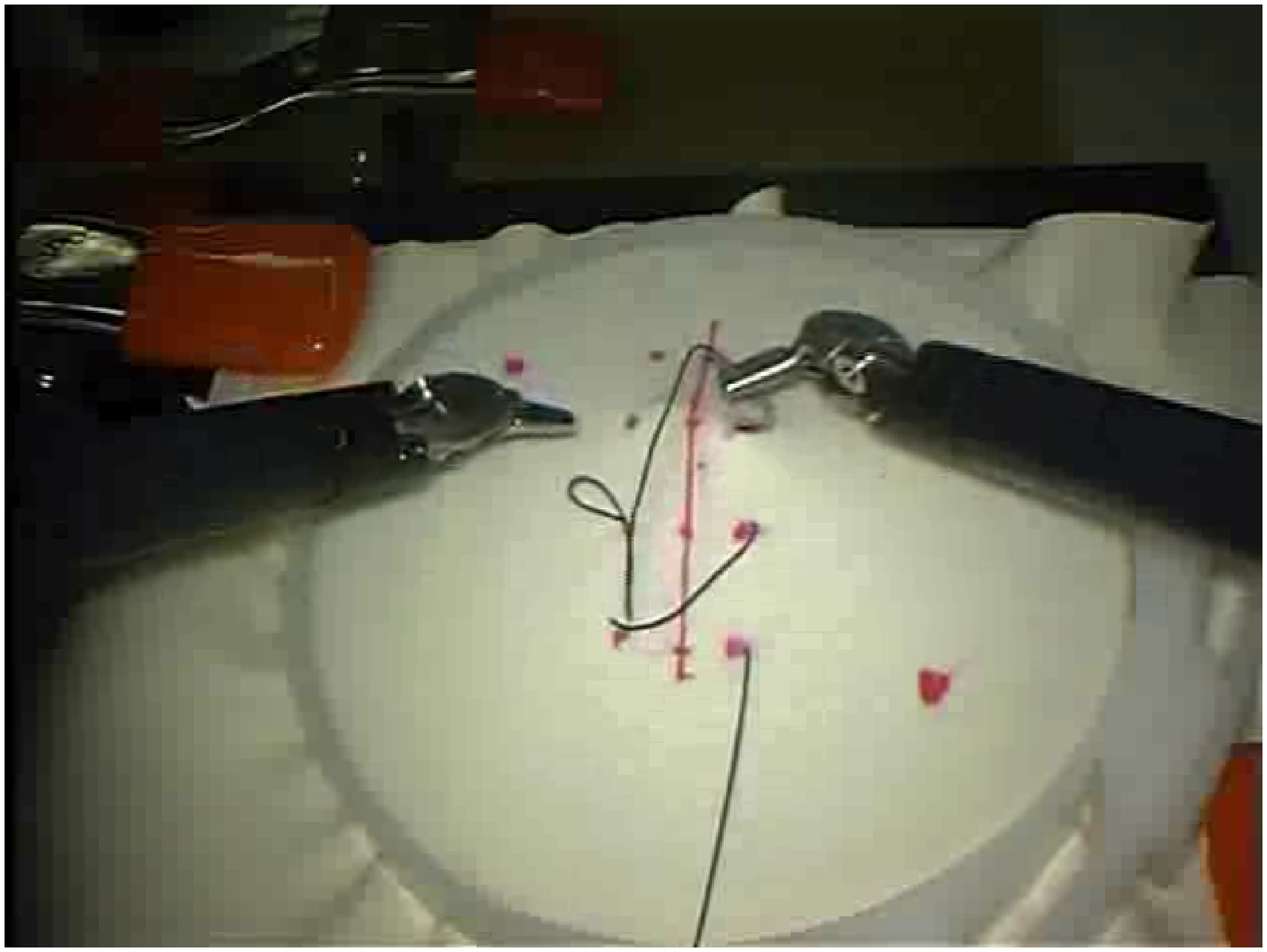}}
  \caption{Surgical operations in JIGSAW dataset: knot tying, needle passing, and suturing.}
  \label{fig:jigsaw}
\end{figure}

We apply the proposed method to surgical robot operators in JIGSAW data set \cite{irl::jigsaw}.
This data set describes three tasks, knot tying, needling passing, and suturing. An illustration of
the tasks is shown in Figure \ref{fig:jigsaw}. Each task is conducted by multiple robot operators,
whose skills range from expert, intermediate to novice.

The data includes videos from two stereo cameras and robot states synchronized to the images. We
assume the operator's actions change the linear and angular acceleration of the robot, and then we
use k-means clustering to identify 10000 actions from the dataset. The state set includes the
robot manipulator's positions and velocities, represented by a length-38 vector with continuous
values. The transition probability is computed based on physical law. 

\begin{figure}
  \centering
  \includegraphics[width=0.4\textwidth]{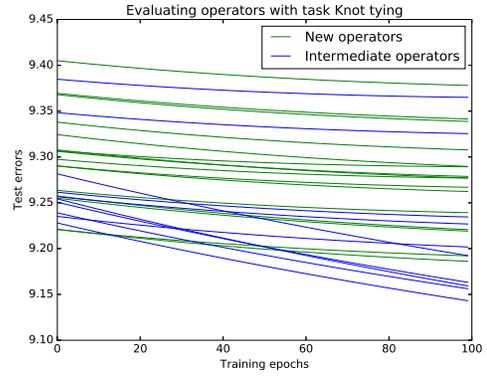}
  \caption{Evaluation of operators on knot tying tasks: "new operators" represent inexperienced
  operators, while "intermediate operators" represent experienced operators. As the training epochs
  increase, the test error decreases, while experienced operators have a relatively lower error rate.}
  \label{fig:knotresult}
\end{figure}

\begin{figure}
  \centering
  \includegraphics[width=0.4\textwidth]{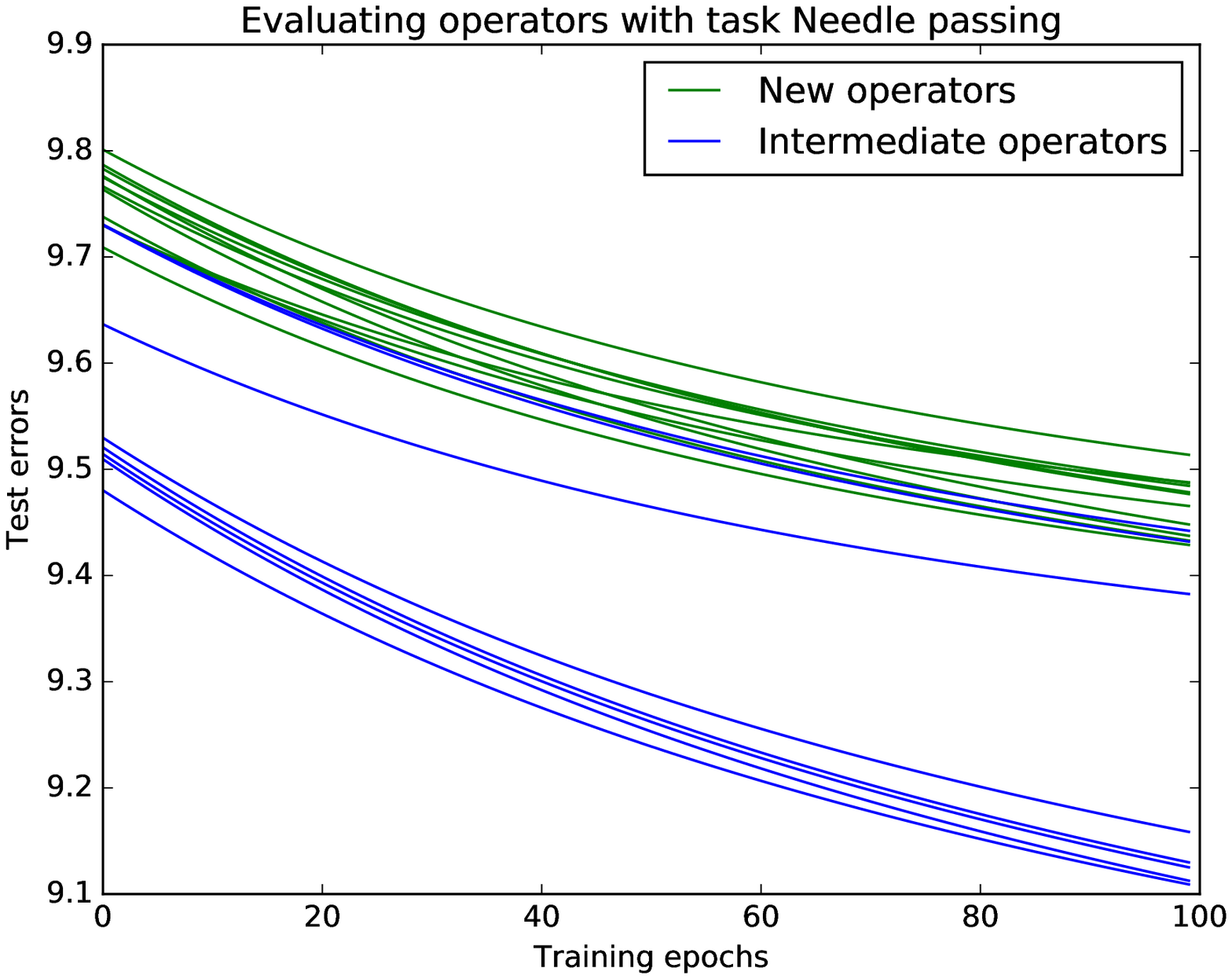}
  \caption{Evaluation of operators on needle passing tasks: "new operators" represent inexperienced
  operators, while "intermediate operators" represent experienced operators. As the training epochs
  increase, the test error decreases, while experienced operators have a relatively lower error rate.}
  \label{fig:needleresult}
\end{figure}

\begin{figure}
  \centering
  \includegraphics[width=0.4\textwidth]{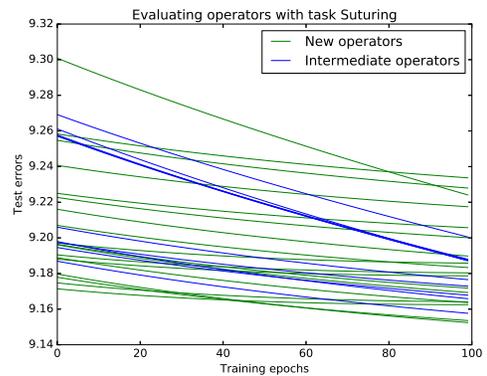}
  \caption{Evaluation of operators on suturing tasks: "new operators" represent inexperienced
  operators, while "intermediate operators" represent experienced operators. As the training epochs
  increase, the test error decreases, while experienced operators have a relatively lower error rate.}
  \label{fig:sutresult}
\end{figure}

We apply the model to surgical operator evaluation on three tasks by training on all experts and
testing on novice and intermediate operators. The results are shown in Figure \ref{fig:knotresult},
\ref{fig:needleresult} and \ref{fig:sutresult}.

The results show that the proposed method successfully identifies the difference between
inexperienced operators and experienced operators, thus it can be used in evaluation tasks.

\section{Conclusions}
\label{irl::conclusions}
This work deals with the problem of high-dimensional inverse reinforcement learning, where the state
space is usually too large for many existing solutions. We solve the problem with a function
approximation framework by approximating the reinforcement learning solution. The method is firstly
tested in a simulated environment, and then applied to the evaluation of surgical robot operators in
three clinical tasks.

In current settings, each task has one reward function, associated with an optimal value function.
In future work, we will extend this method for a robot to learn multiple reward functions. Besides,
we will try to integrate transition model learning into the framework.


\end{document}